\documentclass[11pt]{article}

\usepackage[preprint]{acl}

\usepackage{times}
\usepackage{latexsym}

\usepackage[T1]{fontenc}

\usepackage[utf8]{inputenc}

\usepackage{microtype}

\usepackage{inconsolata}

\usepackage{graphicx}

\usepackage{amsmath}
\usepackage{booktabs}
\usepackage{multirow}
\usepackage{siunitx}
\usepackage{tcolorbox}
\newtcolorbox{mybox}[1]{colback=gray!20,colframe=gray!50!black,fonttitle=\bfseries,title=#1}

\title{Towards Better RL Training Data Utilization via Second-Order Rollout}

\author{Zhe Yang$^{1,2}$, Yudong Wang$^1$, Rang Li$^1$, Zhifang Sui$^1$ \\
$^1$State Key Laboratory of Multimedia Information Processing, \\
School of Computer Science, Peking University \\ 
$^2$ByteDance BandAI
}

\begin{document}
\maketitle
\begin{abstract}

Reinforcement Learning (RL) has empowered Large Language Models (LLMs) with strong reasoning capabilities, but vanilla RL mainly focuses on generation capability improvement by training with only first-order rollout (generating multiple responses for a question), and we argue that this approach fails to fully exploit the potential of training data because of the neglect of critique capability training.
To tackle this problem, we further introduce the concept of second-order rollout (generating multiple critiques for a response) and propose a unified framework for jointly training generation and critique capabilities. 
Extensive experiments across various models and datasets demonstrate that our approach can utilize training data more effectively than vanilla RL and achieve better performance under the same training data. 
Additionally, we uncover several insightful findings regarding second‑order rollout and critique training, such as the importance of label balance in critique training and the noise problem of outcome‑based rewards, which can be mitigated through sampling techniques. 
Our work offers a preliminary exploration of dynamic data augmentation and joint generation‑critique training in RL, providing meaningful inspiration for the further advancement of RL training.

\end{abstract}

\section{Introduction}

\begin{figure}[!t]
    \centering
    \includegraphics[width=0.49\textwidth]{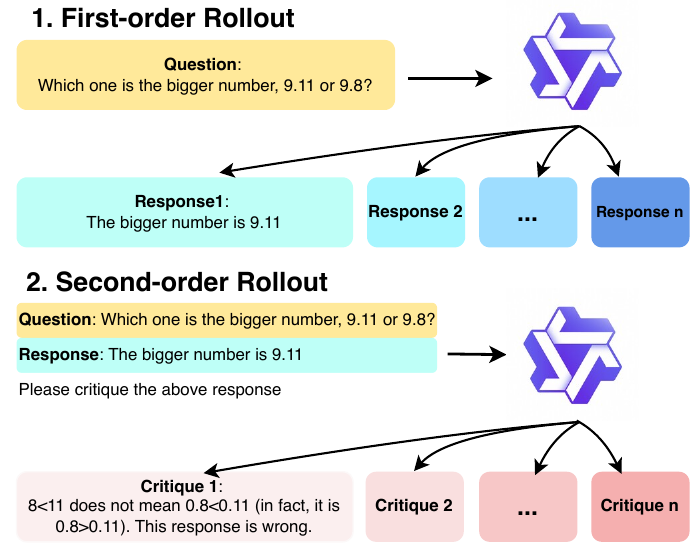}
    \caption{A demonstration of first/second-order rollout. The policy model generates multiple responses for a question in first-order rollout, and generates multiple critiques for a response in second-order rollout.}
    \label{fig:demo}
\end{figure}

With the widespread application of Reinforcement Learning (RL) in post-training \citep{guo2025deepseek}, Large Language Models (LLMs) have demonstrated remarkable reasoning capabilities, which inspires deeper investigations into RL training for LLMs. 
However, current RL training predominantly focuses on enhancing generation capability, often neglecting the development of critique capability, which can be a performance bottleneck for further improvement. 
\citet{Self-critiquing} also categorizes model capabilities into \textit{Generation} and \textit{Critique} \footnote{The original classification delineates three distinct capabilities: Generation, Discrimination, and Critique. For simplicity, we subsume both Discrimination and Critique under the single term Critique.}: 
(1) \textit{Generation} refers to the ability to produce a correct response to a given question; (2) \textit{Critique} denotes the capacity to judge whether a response is correct and to identify specific errors in a wrong response. 
Intuitively, these two capabilities are not independent: \citet{wang2025critique} finds that fine-tuning a model using only critique data, even without any explicit generation data, can significantly improve its generation performance. 
Similarly, better critique performance with only generation training is also witnessed by \citet{wang2025llava}.
The neglect of critique training \citep{yu-etal-2025-training,xie2025teaching} makes us wonder whether current RL training solely on generation capability can fully exploit the potential of training data, and we would like to further explore a joint RL training framework \citep{ruan2025critique,wang2025solving} for better data utilization.


In \S \ref{sec:method} we propose \textbf{G}eneration and \textbf{C}ritique RL (\textbf{GC-RL}) that jointly trains two capabilities with only generation training data by introducing the concept of \textit{second-order rollout}.
In vanilla RL, a policy model samples multiple responses for a given question during training, which we define as \textit{first-order rollout}. Building on this, we further define \textit{second-order rollout} as the process in which the policy model generates multiple critiques for a <question, response> pair, and these two processes are illustrated in Figure \ref{fig:demo}. 
At each training step, we sample questions from the training set for first-order rollout and <question, response> pairs from a data cache for second-order rollout, and these rollout is then combined to update the policy model collectively. 
Meanwhile, responses generated from first-order rollout are filtered and added to the data cache for subsequent training.
It is worth noting that the second-order rollout can proceed naturally based on the results of the first-order rollout, and no additional training data is required, which can be viewed as a "free lunch" to some extent.

In \S \ref{sec:experiment}, extensive experiments are conducted across different models and datasets, demonstrating that our GC-RL shows better data utilization and outperforms vanilla RL in both generation and critique capabilities under the same training data.
Further we conduct more experiments to explore second-order rollout and critique training in \S \ref{sec:analysis} and find: 
1. the data filter is crucial for maintaining balanced critique training (\S \ref{subsec:balanced});
2. outcome-based reward is noisy for critique training and denoising can be achieved through multiple samplings (\S \ref{subsec:denoise});
3. static data performs better in critique-only training and dynamic data are more suitable for joint training (\S \ref{subsec:static});
4. fine-grained model critique behavior manipulation can be achieved through reward function adjustment. (\S \ref{subsec:more_cirtique_reward}).


Our contributions can be summarized as follows:

\begin{enumerate}
    \item We introduce the concept of \textit{second-order rollout} and propose GC-RL framework, which achieves better RL training data utilization by generation-critique joint training.
    \item We conduct extensive experiments to show the effectiveness of our approach and draw some instructive conclusions about critique training.
    \item Our work offers a preliminary exploration of dynamic data augmentation in RL training, providing meaningful inspiration for further advancement of RL.
\end{enumerate}

\section{Related Work}

\paragraph{RL for LLMs}
The exceptional reasoning capabilities demonstrated by Deepseek-R1\citep{guo2025deepseek} highlight the significant role of RL\citep{zhang2025survey} with verifiable reward in training LLMs. 
Beyond rule-based rewards, other forms such as model-based \citep{xu2025tinyv,shao2025deepseekmath} and rubric-based \citep{gunjal2025rubrics,huang2025reinforcement} rewards can also be leveraged for RL training of LLMs. 
Another line of research focuses on developing RL algorithms suited for LLM training, including works like PPO \citep{schulman2017proximal}, GRPO \citep{shao2024deepseekmath}, DAPO \citep{yu2025dapo}, and GSPO \citep{zheng2025group}. 
There are also works exploring new RL training tasks and objectives: for instance, \citet{she2025dupo} employs RL to train models in reconstructing questions from responses, while \citet{dong2025reinforcement} applies RL to enhance next-token prediction. 
Different from previous works, our work explores better RL training utilization through joint training of generation and critique capabilities.

\paragraph{LLM Critique}
The ability to provide critique constitutes a crucial component of LLM capabilities. High-quality critiques enable LLMs to perform self-correction \citep{pan-etal-2024-automatically,yang-etal-2025-confidence,yang-etal-2025-probabilistic} more effectively, and can also enhance the reward signals produced by reward models \citep{yu-etal-2025-self,ankner2024critique,ye-etal-2025-improving} when incorporated into the context. 
\citet{sun-etal-2024-critique} proposes a framework for evaluating the quality of critiques, while other research efforts have focused on improving critique abilities through Supervised Fine-Tuning \citep{wang2025critique}, Direct Preference Optimization \citep{yu-etal-2025-training} and RL \citep{xi2025critique,xie2025teaching,tang2025refcritic}. 
In contrast to prior work, our approach integrates critique training into the vanilla RL framework.

\begin{figure*}[!tb]
    \centering
    \includegraphics[width=0.95\textwidth]{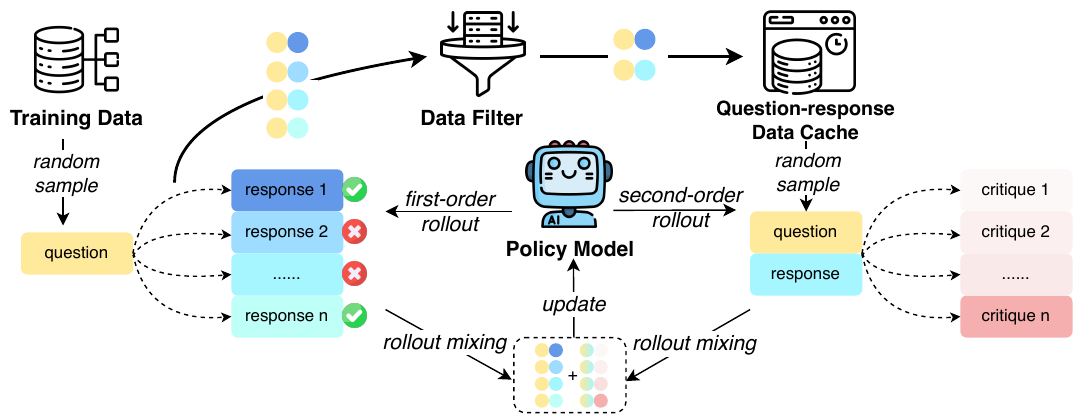}
    \caption{Flowchart of a single training step in GC-RL. First, a batch of questions is sampled from the training data, and multiple responses are generated (first-order rollout). Then, without replacement, a batch of <question, response> pairs is sampled from the \textit{Question-Response Data Cache}, and multiple critiques are generated (second-order rollout). These rollouts are combined and utilized jointly to update the policy model. In addition, the \textit{Question-Response Data Cache} is maintained by processing the first-order rollout through a \textit{Data Filter} and adding the filtered data into the cache.}
    \label{fig:pipeline}
\end{figure*}

\paragraph{Data Augmentation}
Data augmentation \citep{wang2024survey,chai2026text} is an effective approach to enhance model performance in LLM training. Techniques such as back translation \citep{sennrich-etal-2016-improving,kulhanek-etal-2021-augpt} and rephrasing \citep{lu2023epa} can enrich the dataset without altering semantic meanings. Alternatively, another augmentation strategy involves leveraging LLMs to generate new data in zero-shot \citep{oh2023data,ubani2023zeroshotdataaug} or in-context learning \citep{dai2025auggpt} settings. Unlike conventional offline data augmentation methods, our work can be viewed as an online data augmentation process conducted in RL.


\section{Methodology}

\label{sec:method}



We introduce \textbf{G}eneration and \textbf{C}ritique \textbf{RL} (\textbf{GC-RL}), an RL framework for jointly training the generation and critique capabilities, and the overview of our approach is illustrated in the figure \ref{fig:pipeline}.
At each RL training step, a batch of data is sampled from the initial training set, and the policy model generates multiple responses for each question (in conventional RL training, these responses would be directly used to update the policy model). 
We go further by feeding these <question, response> pairs into a \textit{data filter}, retaining a subset to be stored in a \textit{question-response data cache}. 
Then a batch of <question, response> pairs is sampled from the data cache, for which the policy model further generates critiques. 
The responses and critiques obtained from these two rollout processes are combined, assigned rewards and advantages, and utilized to update the policy model.

\paragraph{RL with Second-order Rollout}



For each question $ q_i $ in the training set $ D $, the policy model samples $ n $ responses $ r_1, r_2, \dots, r_n $ during vanilla RL training, a process we refer to as \textit{first-order rollout}. For a specific <question, response> pair $ \langle q_i, r_i \rangle $, the policy model further samples $ n $ critiques $ c_1, c_2, \dots, c_n $ for subsequent training, which we define as \textit{second-order rollout}. 
In essence, the first-order rollout aims to enhance the generation capability, enabling model to produce appropriate answers to given questions. The second-order rollout is designed to improve critique capability, empowering it to identify potential issues or errors within a response.
It is worth noting that the second-order rollout can proceed naturally based on the results of the first-order rollout, and no additional training data is required, which can be viewed as a "free lunch" to some extent.

\paragraph{Critique Data Filter}



During the RL training process, the <question, response> pairs obtained from the first-order rollout are passed through a \textit{Data Filter}, which controls which pairs to retain. 
For a question $q$ and its corresponding responses $r_1, r_2, \dots, r_n$, if all $n$ responses are either entirely correct or entirely incorrect, all these data are discarded. Otherwise, one correct response $ r_{\text{correct}} $ and one incorrect response $ r_{wrong} $ will be randomly selected, and these two resulting pairs $ \langle q, r_{correct} \rangle $ and $ \langle q, r_{wrong} \rangle $ will then stored in the data cache.
Although simple, this data filter plays a crucial role in training stability, and removing this filter would introduce significant issues: 
(1). \textit{imbalance between critique data and generation data}: for a single question $q$, the first-order rollout produces $n$ responses and the second-order rollout further generates $n^2$ critiques, and critique data outnumber generation data by a factor of $n$;
(2). \textit{imbalance within critique data}: we have observed that incorrect responses tend to dominate over correct ones in the first-order rollout, which can cause a label imbalance problem for further critique training, and further discussion is provided in \S \ref{subsec:balanced}.

\paragraph{Mixed Training}
During the update phase of the policy model, we perform training with a mixture of first-order rollout (responses) and second-order rollout (critiques). For each response $r$, a rule-based verifier is employed to check its correctness, with the reward function defined as:
\begin{equation}
R(r) = 
\begin{cases}
1, & \textit{r is correct} \\
0, & \textit{r is wrong}
\end{cases}
\end{equation}
For each critique $c$ generated based on <question,response> pair $\langle q, r \rangle$, the final judgment regarding the correctness of $r$ is extracted, denoted as $\textit{Ext}(c) \in \{\textit{correct}, \textit{wrong}\}$. 
Since intermediate steps of a critique are hard to verify \citep{sun-etal-2024-critique}, we only assign an outcome reward based on the final binary judgment, and the corresponding reward function is:
\begin{equation}
\label{equ:critique_reward}
R(c) = 
\begin{cases}
0.7, & \textit{Ext(c) = correct \& R(r)=1} \\
0.7, & \textit{Ext(c) = wrong \& R(r)=0} \\
0, & else 
\end{cases}
\end{equation}
For responses and critiques in the mixed rollout, the respective rewards are computed with the above reward functions. Then they are mixed into the same group to acquire advantages with GRPO \citep{shao2024deepseekmath} algorithm for further update of policy model.


\section{Experiments}

\label{sec:experiment}

\subsection{Experimental Setup}
\paragraph{Models}
Our experiments are primarily conducted on the Qwen2.5 series \citep{yang2024qwen2}, including Qwen2.5‑(1.5B, 3B, 7B)‑Base, to demonstrate the effectiveness of our method across models of varying scales. Additional experiments are performed on Llama-3.1-8B-Instruct \citep{grattafiori2024llama} and Mistral-7B-Instruct-v0.3 \citep{rastogi2025magistral} to further validate the general applicability of our approach to different model architectures.

\paragraph{Dataset}
The training and evaluation mainly focus on mathematical reasoning tasks. DAPO‑MATH‑17k \citep{yu2025dapo} is employed as the primary training dataset: 1k data are randomly selected to construct cold‑start data, while the remaining 16k data are utilized for RL training. 
Several established mathematical reasoning benchmarks are utilized for evaluation, including Math‑500\citep{math500}, GSM8k\citep{gsm8k}, Minerva\citep{minerva}, AMC23, and OlympiadBench\citep{olympiadbench}.

\paragraph{Implementation Details}
We employ the verl \footnote{\url{https://github.com/volcengine/verl}} framework for RL training and adopt GRPO \citep{shao2024deepseekmath} algorithm, with training hyper‑parameters detailed in the Appendix \ref{app:details}. 


\subsection{RL Training and Experimental Results}

\paragraph{Cold Start}
Applying RL directly to models can lead to several issues:  
(1). Following the format of CFT \citep{wang2025critique}, we instruct the model to append a correctness judgment of its response at the end of the critique. However, the base models exhibit weak instruction‑following capability, often failing to generate critiques that adhere to the required format. 
(2). Due to their limited reasoning performance, the intermediate reasoning steps within the generated critiques are often of low quality. 
To address these problems, we first distill 1,885 initial critique data from GPT-5 \footnote{gpt-5-chat-2025-08-07}, and the prompt used for critique distillation is provided in the Appendix \ref{app:details}. After filtering out data with incorrect formatting or erroneous final judgments, we obtain 1,339 high‑quality training examples. Before starting RL, we perform Supervised Fine‑Tuning (SFT) on this curated critique dataset to equip the model with a preliminary critique capability.

\begin{table*}[!t]
  \centering
  \footnotesize
    \begin{tabular}{cl|ccccc|c}
    \toprule
    \multirow{2}{*}{Models} & \multirow{2}{*}{Methods} & Math-500 & GSM8k & Minerva & AMC23 & Olympiad & \multirow{2}{*}{Avg} \\ 
    \cmidrule(lr){3-7} %
    & & \multicolumn{5}{c|}{\textbf{Generation Accuracy (\%)}} & \\
    \midrule
    \multirow{4}[2]{*}{Qwen2.5-7B} & w/o RL & 55.6  & 77.9  & 16.9  & 35.0  & 22.8  & 41.6  \\
          & C-RL  & 65.1  & 83.7  & 19.2  & 47.5  & 26.1  & 48.3  \\
          & G-RL  & 75.4  & 89.7  & \textbf{24.6} & 60.0  & 33.7  & 56.7  \\
          & GC-RL & \textbf{77.6} & \textbf{92.0} & \textbf{24.6} & \textbf{62.5} & \textbf{39.8} & \textbf{59.3} \\
    \midrule
    \multirow{4}[1]{*}{Qwen2.5-3B} & w/o RL & 20.4  & 29.4  & 4.0   & 5.0   & 7.6   & 13.3  \\
          & C-RL  & 30.4  & 50.0  & 4.4   & 12.5  & 11.0  & 21.7  \\
          & G-RL  & 57.8  & 79.5  & 12.9  & 27.5  & 24.5  & 40.4  \\
          & GC-RL & \textbf{61.8} & \textbf{81.4} & \textbf{14.9} & \textbf{32.5} & \textbf{25.2} & \textbf{43.2} \\
    \midrule
    \multirow{4}[0]{*}{Qwen2.5-1.5B} & w/o RL & 11.0  & 14.6  & 1.8   & 5.0   & 3.3   & 7.1  \\
          & C-RL  & 21.3  & 25.6  & 3.6   & 10.0  & 5.9   & 13.3  \\
          & G-RL  & 45.1  & 67.2  & 8.7   & 25.0  & 12.7  & 31.7  \\
          & GC-RL & \textbf{47.2} & \textbf{69.0} & \textbf{10.3} & \textbf{27.5} & \textbf{15.3} & \textbf{33.9} \\
    \midrule \midrule
    &  & \multicolumn{5}{c|}{\textbf{Critique Accuracy (\%)}} &  \\ 

    \midrule
    \multirow{2}[2]{*}{Qwen2.5-7B} & C-RL  & 80.5  & 82.5  & 62.3  & 71.4  & 72.5  & 73.8  \\
          & GC-RL & \textbf{84.6}  & \textbf{88.3}  & \textbf{67.1}  & \textbf{79.4}  & \textbf{73.8}  & \textbf{78.6} \\
    \midrule
    \multirow{2}[2]{*}{Qwen2.5-3B} & C-RL  & 67.4  & 66.7  & \textbf{57.1} & 64.7  & 61.2  & 63.4  \\
          & GC-RL & \textbf{70.2} & \textbf{72.8} & \textbf{57.1} & \textbf{66.2} & \textbf{61.5} & \textbf{65.6} \\
    \midrule
    \multirow{2}[2]{*}{Qwen2.5-1.5B} & C-RL  & 59.5  & 57.7  & 53.8  & 58.8  & 55.0  & 57.0  \\
          & GC-RL & \textbf{61.4} & \textbf{60.6} & \textbf{57.2} & \textbf{61.3} & \textbf{57.5} & \textbf{59.6} \\
    \bottomrule
    \end{tabular}%
      \caption{Generation and critique capabilities evaluation results on Qwen-2.5-(1.5B,3B,7B). GC-RL outperforms all other RL training methods in both generation and critique capabilities.}
  \label{tab:main_result}%
\end{table*}%

\paragraph{Baselines}
In addition to presenting the evaluation results of our \textbf{GC-RL} (Generation and Critique RL) approach, we also provide the outcomes of several baseline methods: (1) after cold start without RL training; (2) \textbf{G-RL} (Generation RL): vanilla RL training that performs only first-order rollout to enhance generation capability; (3) \textbf{C-RL} (Critique RL): for each question, 10 responses are sampled in advance to construct a balanced training set consisting of <question, response> pairs, and only second-order rollout is performed to train critique capability of LLMs. All these RL training are performed under the same training data.

\paragraph{Critique Evaluation}
In addition to generation capability evaluation, we also try to examine the critique capability after RL training. 
To construct the evaluation dataset, we utilize 5 datasets in generation evaluation as seed data, and sample responses with Qwen2.5-(1.5B,7B,72B)-Instruct, respectively (10 responses for each model).
The final answer is required to be enclosed in \textit{boxed\{\}}, and we filter out responses that dissatisfy this format requirement, as well as questions for which all sampled responses are either entirely correct or entirely incorrect. 
From the remaining data, we randomly select one correct response and an incorrect one for each question, discarding all other responses, and this process yields critique evaluation datasets in which the correct and incorrect responses are 1:1.
Since assessing the accuracy of the intermediate reasoning steps within a critique is particularly challenging, we focus solely on evaluating whether the final judgment of critique on the correctness of the response is accurate, which essentially reduces the task to a binary classification. We also report denoised reward (\S \ref{subsec:denoise}) to measure critique capability as supplemental results in Appendix \ref{app:results}

\paragraph{Results}
The evaluation experiments are conducted on both generation and critique capabilities: 
For generation tasks, model-generated answers are verified by comparing them with reference answers and the final accuracy is reported;
For critique tasks, we extract the generated final judgment on the correctness of a response and measure the binary classification accuracy.
We show experimental results for Qwen-2.5-(1.5B,3B,7B) in Table \ref{tab:main_result}, and provide more results on Llama-3.1-8B-instruct and Mistral-7B-Instruct-v0.3 in Appendix \ref{app:results}, finding that: 
1. Even if a model is trained solely with C-RL (without generation training), its generation ability significantly improves—though such enhancement remains far inferior to that achieved through G-RL.
2. After training with GC-RL, the model attains optimal performance in both generation and critique capabilities. On one hand, its generation capability surpasses that of models trained with G-RL; on the other hand, its critique ability exceeds that of models trained with C-RL. This result suggests a certain coupling between critique and generation capabilities, and further demonstrates that joint training yields superior overall performance compared to training for each capability independently.

\section{Analysis}

\begin{table*}[!t]
  \centering
  \footnotesize
    \begin{tabular}{cl|ccccc|c}
    \toprule
          &       & Math-500 & GSM8k & Minerva & AMC23 & Olympiad & Avg \\
    \midrule
    \multirow{3}[2]{*}{Genration} & Random Sampling & 75.0  & 90.9  & 22.9  & 60.0  & 37.5  & 57.3  \\
          & Random Sampling + Reweight & 77.2  & 91.5  & 23.2  & 60.0  & 38.2  & 58.0  \\
          & Data Filter & \textbf{77.6 } & \textbf{92.0 } & \textbf{24.6 } & \textbf{62.5 } & \textbf{39.8 } & \textbf{59.3 } \\
    \midrule
    \multirow{3}[2]{*}{Critique } & Random Sampling & 82.3  & 84.0  & 63.3  & 73.5  & 71.6  & 74.9  \\
          & Random Sampling + Reweight & 83.7  & 86.7  & 65.8  & 77.9  & 72.8  & 77.4  \\
          & Data Filter & \textbf{84.6} & \textbf{88.3} & \textbf{67.1} & \textbf{79.4} & \textbf{73.8} & \textbf{78.6} \\
    \bottomrule
    \end{tabular}%
    \caption{A comparison of model performance on Qwen2.5-7B when sampling responses with/without a data filter. Random sampling leads to the worst performance, though adding reward reweighting can alleviate this issue. Utilizing the data filter achieves the best performance.}
  \label{tab:reweight_result}%
\end{table*}%

\label{sec:analysis}
First-order rollout in RL training has been thoroughly studied by previous works, so we conduct a more detailed analysis of second-order rollout and critique training in this section. 
First, we discuss the label imbalance problem in critique training and demonstrate the effectiveness of our data filter both theoretically and experimentally (\S \ref{subsec:balanced}). 
Next, we discuss the reward noise problem in critique training and explore a sampling-based denoising strategy (\S \ref{subsec:denoise}). 
We then compare static and dynamic data in critique training, observing that dynamic data is more suitable for GC-RL, while static data works better for C-RL (\S \ref{subsec:static}). 
Finally, by adjusting the reward function, we achieve fine-grained critique behavior manipulation of LLMs after RL training (\S \ref{subsec:more_cirtique_reward}).

\subsection{Towards Balanced Critique Training}
\label{subsec:balanced}



We theoretically analyze why GC-RL without a data filter can lead to training data imbalance and restrict the critique capability of LLMs. To mitigate this problem, we explore employing a reward reweighting strategy and utilizing a data filter, and finally conduct comparative experiments to validate the effectiveness of our data filter.
\paragraph{Alleviating imbalance problem with reward reweighting}
The ultimate judgement of whether a response is correct or not in a critique is essentially a binary classification task, but the number of erroneous responses significantly outweighs the correct ones in the first-order rollout.
Intuitively, this label imbalance problem can impact subsequent critique training, and we also theoretically analyze the effect of data imbalance on the performance of the critique training, with detailed discussions shown in Appendix \ref{app:imbalance}.
To mitigate this problem, we also explore reweighting and scaling rewards for positive and negative data to balance their contributions, and the weighted reward function is defined as:
\begin{equation}
\label{equ:weighted_reward_function}
R_{w}(c) = 
\begin{cases}
\frac{0.35}{E[R(r)]}, & \textit{Ext(c) = correct \& R(r)=1} \\
\frac{0.35}{1-E[R(r)]}, & \textit{Ext(c) = wrong \& R(r)=0} \\
0, & else 
\end{cases}
\end{equation}
where $E[R(r)]$ is the expected reward for response $r$ during RL training.
It can be mathematically proved that the weighted reward is unbiased and does not incentivize the model to judge an uncertain response as correct or wrong, and the detailed proof is shown in Appendix \ref{app:imbalance}.
Essentially, this weighting strategy amplifies the reward for rare classes, enabling the model to learn more effectively from such data and thereby approximating balanced training.

\paragraph{Empirical comparison of training with/without a data filter}
Comparison experiments are conducted on Qwen2.5-7B with GC-RL training under three settings: 
(1) randomly sampling responses without data filter, (2) randomly sampling responses and training with the weighted reward function in Equation \ref{equ:weighted_reward_function}, and (3) utilizing the data filter in \S \ref{sec:method}.
As the experimental results shown in Table \ref{tab:reweight_result}, random sampling strategy leads to the worst generation and critique capabilities because of label imbalance, and this problem can be alleviated to some extent with a weighted reward function. 
Although reward weighting can achieve balance in the reward level for imbalanced data, utilizing a data filter can achieve inherently balanced training data and yield the best performance.

\subsection{Reward Noise \& Denoising in Critique RL}
\label{subsec:denoise}




\paragraph{Reward noise problem in critique RL}
Obtaining precise rewards for critiques is more challenging than for responses. For instance, in mathematical problems, since we have a corresponding answer to each question, the correctness of a response can be easily rule-based verified and a precise reward can then be assigned accordingly.
However, for critiques, it is difficult to verify the correctness of each intermediate step, and we can only assign rewards based on whether the final binary classification result is correct. 
Generating responses is essentially a generation task, and it is rare for a model to produce intermediate errors yet still arrive at the correct final answer. In contrast, generating critiques is essentially a binary classification task, and even random guessing can yield correct answers with a 50\% probability, leading to many critiques with incorrect intermediate steps but correct final judgement. 
Ideally, for critiques with correct outcomes, we should differentiate between those with erroneous intermediate steps and those with correct ones, assigning lower and higher rewards, respectively.
However, verifying intermediate steps is challenging in practice, so critiques with correct results are often given the same reward, and we refer to such rewards as \textit{noised rewards}.
\citet{guo2025deepseek} has demonstrated significant success in Reinforcement Learning with Verifiable Rewards (RLVR), which fundamentally relies on \textit{oracle rewards}. \citet{liu2504inference,shi2025heimdall,whitehouse2025j1} also attempt RL on classification tasks, showing that even with \textit{noised rewards}, model performance can be improved to some extent, which is also witnessed by our experiments in \S \ref{sec:experiment}.
\begin{figure}[!t]
    \centering
    \includegraphics[width=0.47\textwidth]{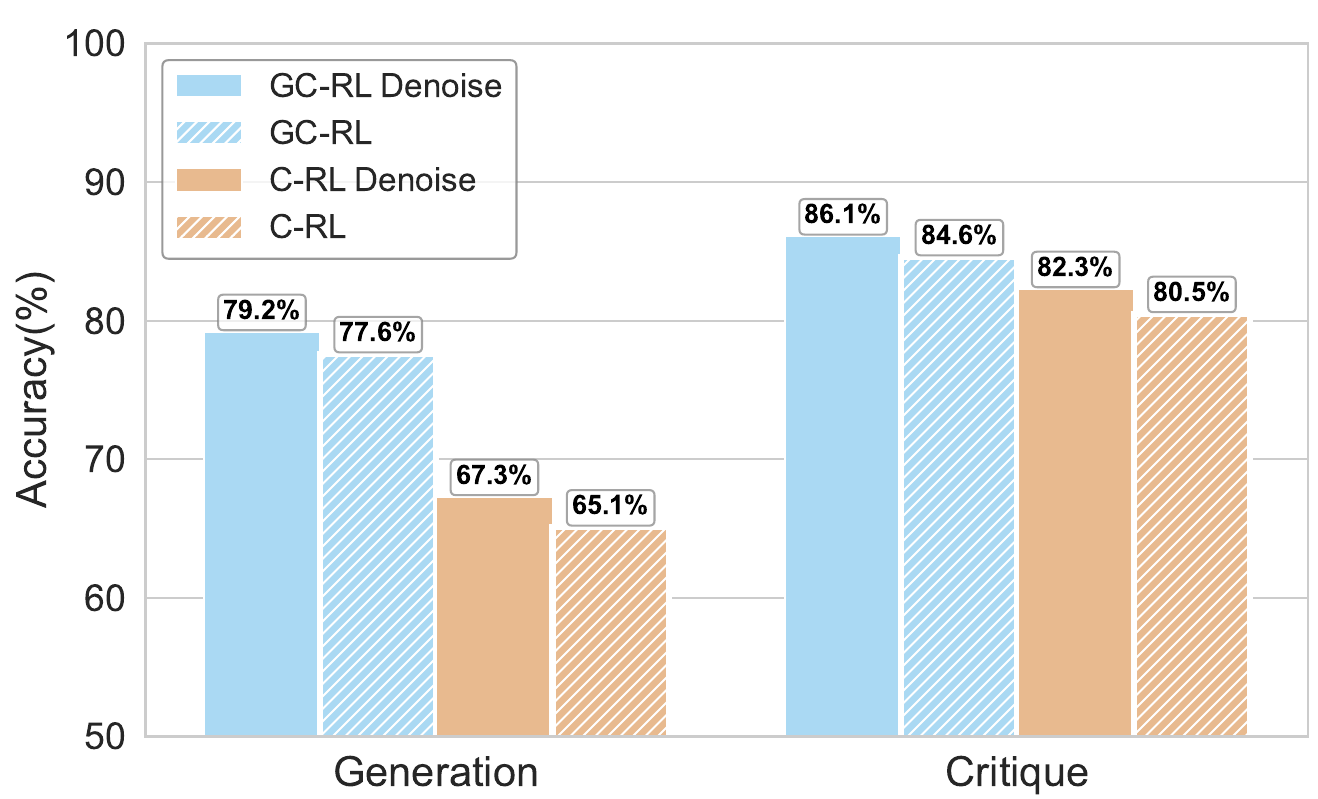}
    \caption{A comparison of model performance of Qwen2.5-7B with/without reward denoising strategy on Math-500. In both GC-RL and C-RL settings, reward denoising can improve model performance on both generation and critique capabilities.}
    \label{fig:bar_denoise}
\end{figure}
%


\paragraph{Exploring critique reward denoising based on self-correction with multiple samplings}
In LLM self-correction process \citep{kamoi-etal-2024-llms,pan-etal-2024-automatically}, a model is provided with three components: <question, response, critique>, and then instructed to modify the original response based on the potential issues identified by the critique and generate a refined response.
Intuitively, the higher the quality of the critique—i.e., the more accurately it identifies problems in the original response—the greater the likelihood that the refined response will be correct. 
Thus, we can inversely estimate the quality of a critique based on the quality of its corresponding refined response. 
Similar to \citep{tang2025refcritic,xie2025teaching,yu-etal-2025-training}, for a given critique, we allow the model to perform self-correction and sample \(n\) refined responses, with the number of correct ones denoted as \(k\), based on which we then propose the following reward function to estimate critique quality:
\begin{equation}
\label{equ:reward_est}
R_q(c) = 
\begin{cases}
0.1*\frac{k}{n}, & \textit{Ext(c) == correct} \\
0.7*\frac{k}{n}, & \textit{Ext(c) == wrong}
\end{cases}
\end{equation}
Although it is difficult to directly verify the correctness of intermediate steps in critique, this sampling method can indirectly estimate it to some extent, thereby reducing noise in the reward. 
We utilize the sum of the outcome reward from Equation \ref{equ:critique_reward} and the estimated reward obtained from Equation \ref{equ:reward_est}, denoted as $R(c) + R_q(c)$, as the final reward for critique, and compare it with using only the outcome reward $R(c)$. 
Theoretically, a larger number of samples $n$ leads to better noise reduction and more accurate reward values, but at a higher computational cost. To make the computational overhead controllable, experiments are conducted with $n = 1$, and the experimental results on Math-500 are shown in Figure \ref{fig:bar_denoise} (with more results shown in Figure \ref{fig:bar_denoise_22} in Appendix \ref{app:results}). A performance improvement in both generation and critique capabilities can be witnessed under both CG-RL and C-RL settings with our reward denoising strategy. 

\subsection{Static v.s. Dynamic Data}
\label{subsec:static}
\paragraph{A comparison of static and dynamic responses during critique RL training.}
In our approach, dynamic self-generated responses are utilized when generating second-order rollout, and an alternative strategy involves utilizing static responses \citep{ruan2025critique,xie2025teaching} for critique RL training, where pre-prepared <question, response> pairs remain fixed throughout the RL process. 
A performance comparison of static and dynamic training data is conducted under GC-RL and C-RL settings for both generation and critique capabilities on Qwen2.5-7B, and the experimental results on Math-500 are presented in Figure \ref{fig:bar_static} (with more results shown in Figure \ref{fig:bar_static_22} in Appendix \ref{app:results}).
We find that training with dynamically self-generated response data leads to higher performance in both generation and critique capabilities under GC-RL setting.  
However, under C-RL setting, where responses are abandoned and only critiques are utilized to update the policy model, dynamic data suffers from a severe reward hacking problem, and the static data strategy significantly outperforms the dynamic data. 
With dynamic data, the model seems to identify a shortcut to maximize rewards during RL: it deliberately generates incorrect responses in the generation stage (though producing a correct response is challenging, generating an incorrect one can be quite easy), then it labels all responses as incorrect to get the reward in the critique stage.
To summarize, dynamic data is more suitable for GC-RL training, while static data is more appropriate for C-RL training.
\begin{figure}[!t]
    \centering
    \includegraphics[width=0.47\textwidth]{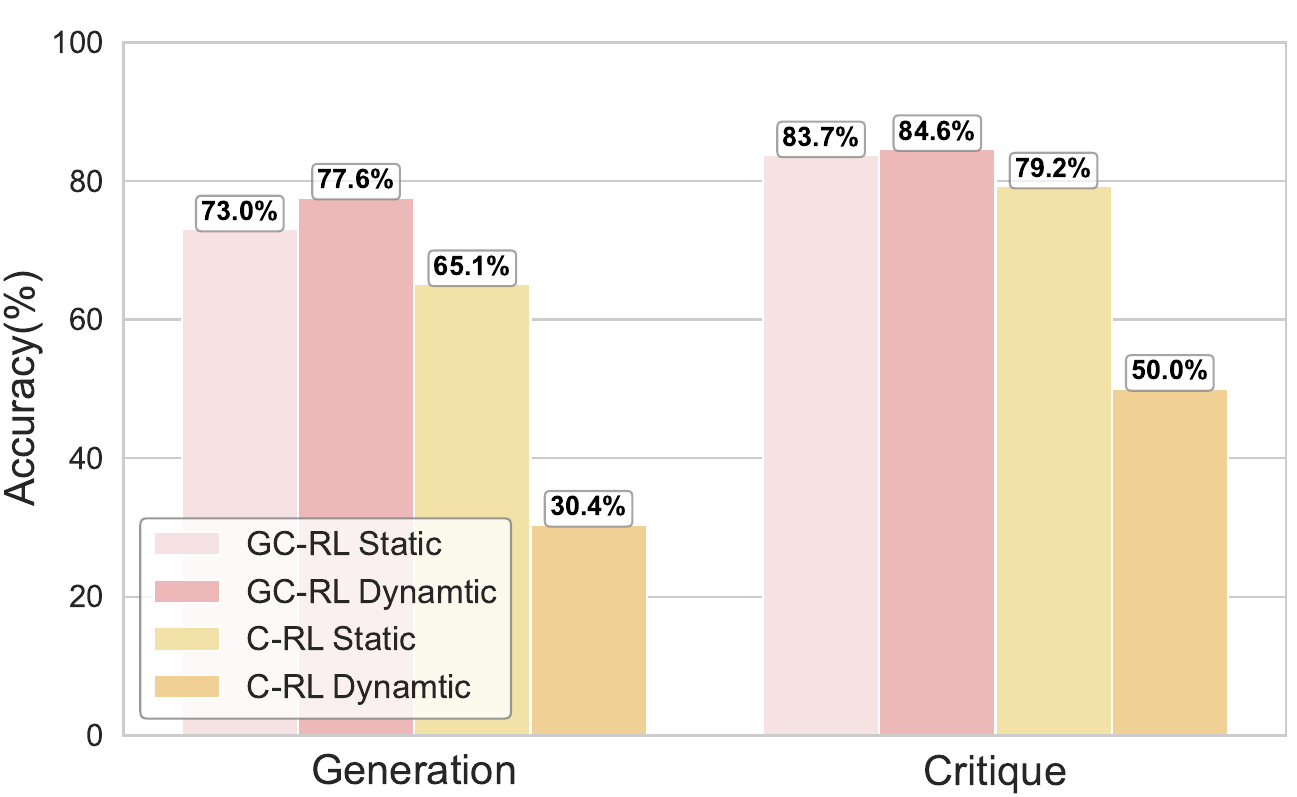}
    \caption{Performance of Qwen2.5-7B on Math-500 under GC-RL and C-RL settings with static and dynamic critique training data. Dynamic data outperforms static data in the GC-RL setting, while the opposite holds for C-RL.}
    \label{fig:bar_static}
\end{figure}

\subsection{Fine-grained Critique Behavior Manipulation}
\label{subsec:more_cirtique_reward}

There are often different requirements for the binary classification performance in different scenarios. For instance, a higher recall rate is demanded in disease screening, while a higher precision rate is required in recommendation systems.
In the reward function defined in Equation \ref{equ:critique_reward}, the same reward is assigned to both correct and incorrect responses as long as the critique identifies them correctly, and we also explore training with more fine-grained reward functions to better control the critique behavior of LLMs. For example, to encourage the model to be more inclined to classify a response as incorrect when it is uncertain, we try the following reward function:
\begin{equation}
R_w(c) = 
\begin{cases}
0.6, & \textit{Ext(c) = correct \& R(r)=1} \\
0.8, & \textit{Ext(c) = wrong \& R(r)=0} \\
0, & else 
\end{cases}
\end{equation}

Conversely, to steer the model toward classifying uncertain responses as correct, we assign a large reward value to correct responses and employ the following reward function:

\begin{equation}
R_r(c) = 
\begin{cases}
0.8, & \textit{Ext(c) = correct \& R(r)=1} \\
0.6, & \textit{Ext(c) = wrong \& R(r)=0} \\
0, & else 
\end{cases}
\end{equation}

The experimental results on Math-500 of utilizing \(R_w(c)\) and \(R_r(c)\) are presented in Figure \ref{fig:bar_reward}, and more results can be found in Figure \ref{fig:bar_reward_22} in Appendix \ref{app:results}). Compared with the baseline \(R(c)\), we observe that when \(R_w(c)\) is applied, the model achieves higher precision but lower recall; whereas with \(R_r(c)\), the precision decreases while recall increases. By adjusting the reward function, we can exert finer-grained control over the critique behavior of LLMs.
\begin{figure}[!t]
    \centering
    \includegraphics[width=0.47\textwidth]{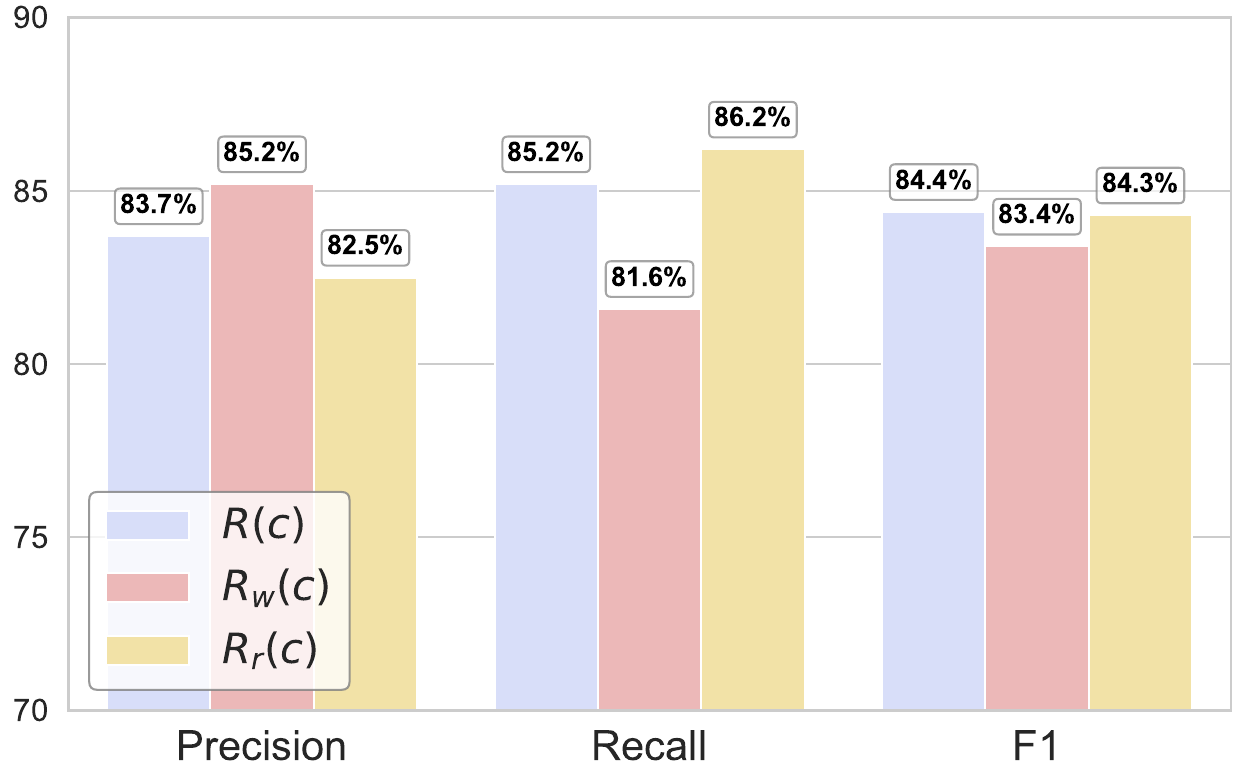}
    \caption{A comparison of critique performance of Qwen2.5-7B with different reward function on Math-500. Compared to baseline $R(c)$, \(R_w(c)\) leads to a higher precision while \(R_r(c)\) generates a higher recall.}
    \label{fig:bar_reward}
\end{figure}


\section{Conclusion}

Based on the first-order rollout in vanilla RL (generating multiple responses for a question), we further introduce the concept of second-order rollout (generating multiple critiques for a response) and propose GC‑RL, a unified framework to train generation and critique capabilities jointly. 
Extensive experiments across various models and datasets demonstrate that our approach can more effectively utilize training data compared to vanilla RL, achieving superior performance under the same training data. 
Additionally, we uncover some insightful findings related to second‑order rollout and critique training, such as the importance of label balance in critique training. 
Our work serves as an initial exploration into dynamic data augmentation and joint training of generation and critique in RL training, offering meaningful insights for further advancements in RL for LLMs.


\section*{Limitations}


Our work represents a preliminary attempt to integrate dynamic data augmentation with joint training of generation and critique into RL, and still exhibits several limitations that warrant further exploration. 
For simplicity, we only employ the GRPO algorithm for RL training, and the applicability of our approach to other RL algorithms (such as PPO) remains to be investigated.
Moreover, experiments are confined to the mathematical domain using models with fewer than 10B parameters, and extending our approach to larger-scale RL training with multi-domain training data and larger models is considered to be an important direction for future work.
Additionally, we have observed that our approach converges more slowly compared to vanilla RL, essentially trading computational resources for improved performance. 
Our approach also requires that responses can be rule-based verified, making it less straightforward to apply to RL tasks where responses are harder to evaluate (e.g., rubric-based RL training \citep{gunjal2025rubrics,huang2025reinforcement}). How to perform second-order rollout and assign rewards to critiques in such settings is also worth further exploration.
\section*{Ethical Considerations}
The data utilized are open for research, and LLMs in the experiments are all publicly available by either parameters or API calls. Therefore, we do not anticipate any ethical concerns in our research.


\bibliography{custom}

\appendix

\section*{Appendix}
\label{sec:appendix}

\section{More Implementation Details}
\label{app:details}
We provide more implementation details in this section. We show the prompt utilized for cold start data distillation in Figure
\ref{fig:prompt}. We show some hyper-parameters in RL training in Table \ref{tab:training_hyper}.

\begin{figure}[!h]

\begin{mybox}{Prompt for Critique Data Distillation}
\footnotesize

\textbf{\#Question\#:}\\
<insert question> \\
\textbf{\#Solution\#:}\\
<insert question> \\
\textbf{\#Instruction\#:}\\
Please verify step by step and judge whether the solution is correct, and end your answer with **Conclusion: right/wrong [END]**

\end{mybox}
\caption{Our prompt fed to GPT-5 for critique data generation, which is comprised of a question, a solution, and critique instruction.}
\label{fig:prompt}
\end{figure}

\begin{table}[!ht]
  \centering
  \footnotesize
    \begin{tabular}{l|c}
    \toprule
    train bath size & 512 \\ \midrule
    ppo mini batch size & 128 \\ \midrule
    rollout n & 5 \\ \midrule
    adv estimator & grpo \\ \midrule
    kl loss coef & 1e-3 \\ \midrule
    learning rate & 1e-6 \\ \midrule
    max prompt length & 4096 \\ \midrule
    max response length & 4096 \\ \midrule
    clip ratio & 0.2 \\ \midrule
    epochs & 10 \\
    \bottomrule
    \end{tabular}
\caption{RL training hyper-parameters.}
\label{tab:training_hyper}
\end{table}

\section{More Experimential Results}
\label{app:results}
\begin{figure*}[!t]
    \centering
    \includegraphics[width=0.95\textwidth]{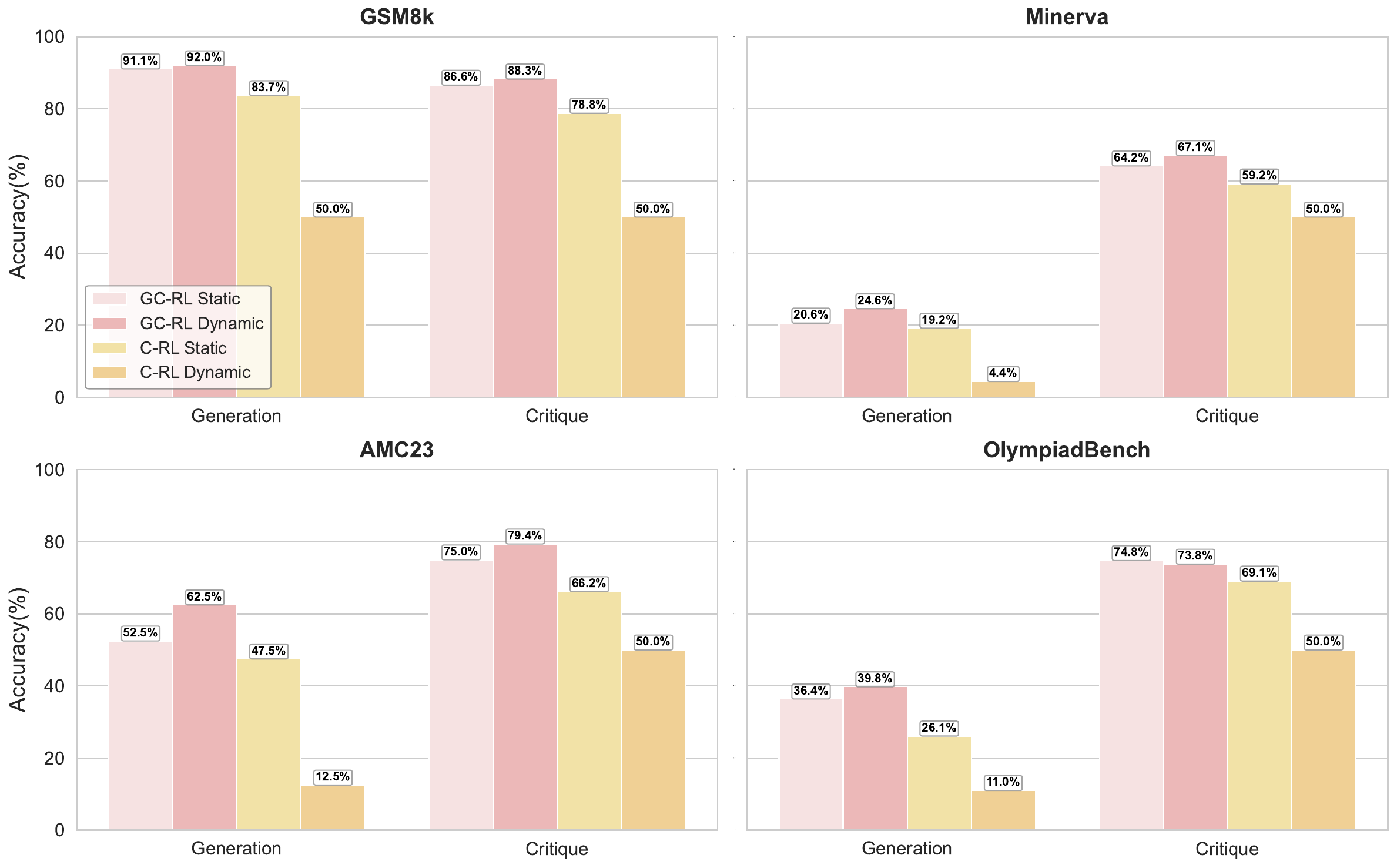}
    \caption{Performance of Qwen2.5-7B on 4 datasets under GC-RL and C-RL settings with static and dynamic critique training data. Dynamic data outperforms static data in the GC-RL setting, while the opposite holds for C-RL.}
    \label{fig:bar_static_22}
\end{figure*}

\begin{figure*}[!t]
    \centering
    \includegraphics[width=0.95\textwidth]{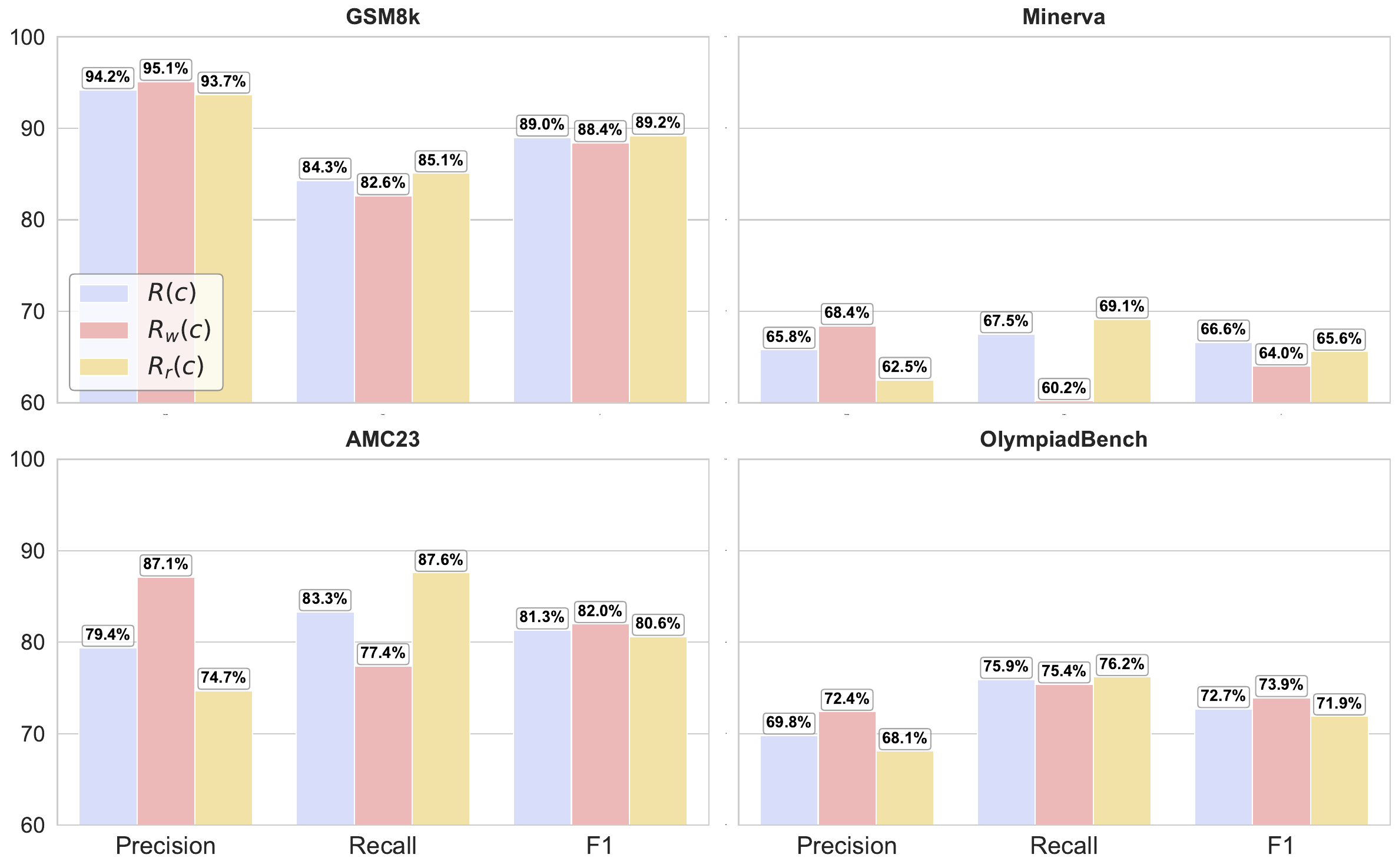}
    \caption{A comparison of critique performance of Qwen2.5-7B with different reward functions on 4 datasets. Compared to baseline $R(c)$, \(R_w(c)\) leads to a higher precision while \(R_r(c)\) generates a higher recall.}
    \label{fig:bar_reward_22}
\end{figure*}

\begin{figure*}[!t]
    \centering
    \includegraphics[width=0.95\textwidth]{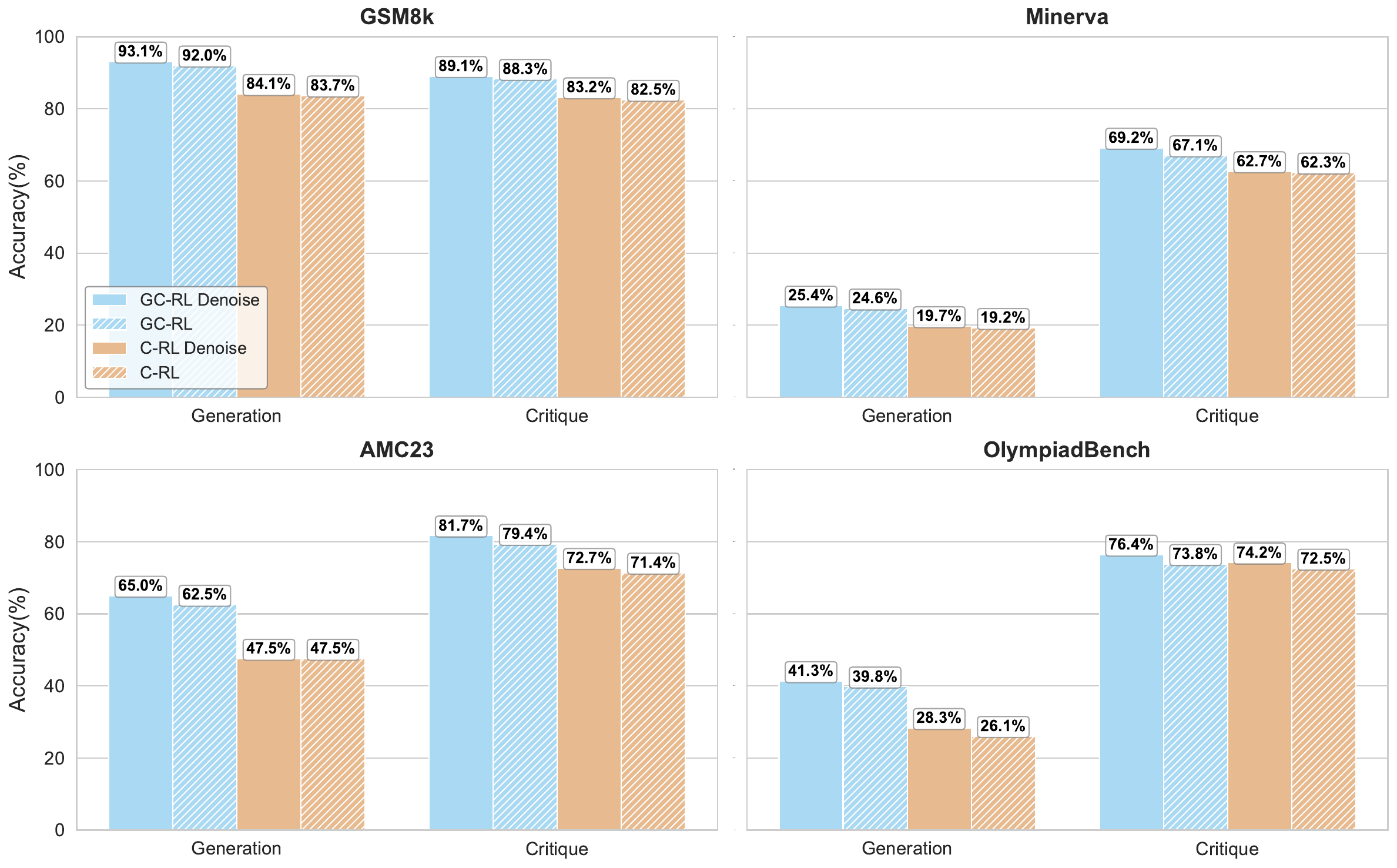}
    \caption{A comparison of model performance of Qwen2.5-7B with/without reward denoising strategy on 4 datasets. In both GC-RL and C-RL settings, reward denoising can improve model performance on both generation and critique capabilities.}
    \label{fig:bar_denoise_22}
\end{figure*}

We show more experimental results in this section. Main experiments on Mistral-7B-Instruct-v0.3 and Llama-3.1-8B-Instruct are shown in Table \ref{tab:main_result_app}. Supplemental critique capability evaluation results are shown in Table \ref{tab:denoised_reward_eval}. Performance comparison of static and dynamic critique training data is shown in Figure \ref{fig:bar_static_22}; performance comparison of different reward functions is shown in Figure \ref{fig:bar_reward_22}; performance comparison of training with/without reward denoising strategy is shown in Figure \ref{fig:bar_denoise_22}.

\section{Discussions on Label Imbalance Problem}
\label{app:imbalance}

\paragraph{A theoretical analysis of data imbalance problem in critique training without the data filter.}
Theoretically, generating a critique can be viewed as a binary classification task, requiring the critique to include a final judgment on whether the response is correct or not, and binary classification tasks are susceptible to label distribution imbalance in the training data: when one category predominates, the trained model tends to favor predicting that category. 
In \S \ref{sec:method}, the data processed through the first-order rollout and subsequently filtered by a data filter are stored in the \textit{Question-response Data Cache} to support subsequent critique RL training, and this data filter ensures a 1:1 ratio between correct and incorrect responses, thereby preventing label bias in subsequent critique training.
If the data filter is omitted and responses are sampled uniformly at random, in the first-order rollout, $E[R(r)] \times 100\%$ of responses are correct and $(1 - E[R(r)]) \times 100\%$ are incorrect, where $E$ denotes the mathematical expectation. Consequently, the ratio of correct to incorrect responses under random sampling is also $E[R(r)] : (1 - E[R(r)])$. 
Similar to \citet{yang-etal-2025-confidence}, let $P_1$ and $P_2$ denote the probabilities that the model correctly identifies correct and wrong responses, respectively, during the critique phase. 
On a validation set with a balanced 1:1 positive-to-negative ratio, the expected validation reward is $E[R_{val}(c)] = \frac{0.7P_1 + 0.7P_2}{2}$, and the expected reward for critiques during RL training is:
\begin{equation}
\label{equ:expected_R(c)}
\begin{aligned}
& \quad E[R(c)] \\
&= E[R(r)]*P_1*0.7 + (1- E[R(r)])*P_2*0.7 \\
&= 0.7E[R(r)]P_1 \\ & \quad + (1- E[R(r)])(2E[R_{val}(c)]-0.7P_1) \\
&= 0.7(2E[R(r)]-1)P_1 \\ & \quad + 2(1- E[R(r)])E[R_{val}(c)]
\end{aligned}
\end{equation}
Though $E[R(r)]$ and $E[R_{val}(c)]$ gradually improve throughout the whole RL training process, they can be approximately viewed as constants between adjacent RL steps.
From this point of view and Equation \ref{equ:expected_R(c)}, we can see that the expected critique reward is in proportion to $P_1$. Similar to \citet{yang-etal-2025-confidence}, $P_1$ and $P_2$ also exhibit a competitive trade-off and $P_1+P_2$ can be viewed as a constant between adjacent RL steps.
When $2E[R(r)] - 1 > 0$, this reward encourages the model to increase $P_1$ and decrease $P_2$; when $2E[R(r)] - 1 < 0$, a decrease in $P_1$ raises the expected reward, thereby incentivizing the model to reduce $P_1$ and increase $P_2$. For example, when training Qwen2.5-7B with GC-RL in \S \ref{sec:experiment}, we find $E[R(r)] \approx 0.1$ in early training and $E[R(r)] \approx 0.45$ in late stage, consistently satisfying $2E[R(r)] - 1 < 0$ and incentivizing the model to decrease $P_1$. 
Consequently, a model trained with randomly sampled responses for second-order rollout exhibits lower $P_1$ and higher $P_2$, meaning it tends to classify responses as incorrect during critique.

\paragraph{The weighted reward function is unbiased}
We give a proof of the unbiasedness of the following weighted reward function: 
\begin{equation}
R_{w}(c) = 
\begin{cases}
\frac{0.35}{E[R(r)]}, & \textit{Ext(c) = correct \& R(r)=1} \\
\frac{0.35}{1-E[R(r)]}, & \textit{Ext(c) = wrong \& R(r)=0} \\
0, & else 
\end{cases}
\end{equation}
Under this scheme, the expected reward for critiques during RL training becomes: 
\begin{equation}
\begin{aligned}
& \quad E[R(c)] \\
&= E[R(r)]P_1 \frac{0.7}{2E[R(r)]} \\ & \quad + (1- E[R(r)])P_2 \frac{0.7}{2(1-E[R(r)])} \\
&= \frac{0.7P_1 + 0.7P_2}{2} \\
&= E[R_{val}(c)]
\end{aligned}
\end{equation}
Under these circumstances, the reward does not incentivize the model to increase or decrease $P_1$ or $P_2$. Essentially, this weighting strategy amplifies the reward for rare classes, enabling the model to learn more effectively from such examples and thereby approximating balanced training.

\begin{table*}[!t]
  \centering
  \footnotesize
    \begin{tabular}{cl|ccccc|c}
    \toprule
    \multirow{2}{*}{Models} & \multirow{2}{*}{Methods} & Math-500 & GSM8k & Minerva & AMC23 & Olympiad & \multirow{2}{*}{Avg} \\ 
    \cmidrule(lr){3-7} %
    & & \multicolumn{5}{c|}{\textbf{Generation Accuracy (\%)}} & \\
    \midrule
    \multirow{4}[2]{*}{Mistral-7B-Instruct-v0.3} & w/o RL & 10.8 & 46.7 & 7.4 & 5.0 & 1.8 & 14.3 \\
    & C-RL  & 19.8  & 54.2  & 9.8   & 12.5  & 4.2 & 20.1 \\
    & G-RL  & 47.6  & 77.8  & 18.6  & 30.0  & 15.6 & 37.9 \\
    & GC-RL & \textbf{52.1}  & \textbf{81.2}  & \textbf{19.4}  & \textbf{32.5}  & \textbf{17.9} & \textbf{40.6}  \\
    \midrule
    \multirow{4}[1]{*}{Llama-3.1-8B-Instruct} & w/o RL & 48.2 & 84.2 & 19.1 & 25.0 & 15.0 & 38.3 \\
    & C-RL & 55.6 & 87.6 & 21.3 & 32.5 & 20.1 & 43.4 \\
    & G-RL & 72.3 & 92.0 & 28.9 & 47.5 & 28.4 & 53.8 \\
    & GC-RL & \textbf{75.8} & \textbf{92.6} & \textbf{30.1} & \textbf{50.0} & \textbf{31.6} & \textbf{56.0} \\
    \midrule \midrule
    &  & \multicolumn{5}{c|}{\textbf{Critique Accuracy (\%)}} &  \\ 
    \midrule
    \multirow{2}[2]{*}{Mistral-7B-Instruct-v0.3} & C-RL  & 65.2 & 70.1 & 60.3 & 66.4 & 63.2 & 65.0  \\
          & GC-RL & \textbf{69.0} & \textbf{74.8} & \textbf{64.5} & \textbf{71.6} & \textbf{66.1} & \textbf{69.2} \\
    \midrule
    \multirow{2}[2]{*}{Llama-3.1-8B-Instruct} & C-RL  & 77.8  & 80.5  & 58.3  & 70.4  & 67.3   & 70.9  \\
          & GC-RL & \textbf{81.4} & \textbf{83.2} & \textbf{64.1} & \textbf{75.4} & \textbf{70.6} & \textbf{74.9} \\
    \bottomrule
    \end{tabular}%
      \caption{Generation and critique capabilities evaluation results on Mistral-7B and Llama3.1-8B-Instruct. GC-RL outperforms all other RL training methods in both generation and critique capabilities.}
  \label{tab:main_result_app}%
\end{table*}%

\begin{table*}[!t]
  \centering
  \footnotesize
    \begin{tabular}{cl|ccccc|c}
    \toprule
    \multirow{2}{*}{Models} & \multirow{2}{*}{Methods} & Math-500 & GSM8k & Minerva & AMC23 & Olympiad & \multirow{2}{*}{Avg} \\ 
    \cmidrule(lr){3-7} %
    & & \multicolumn{5}{c|}{\textbf{Denoised Reward}} & \\
    \midrule
    \multirow{2}[2]{*}{Qwen2.5-7B} & C-RL  & 0.901 & 0.912 & 0.715 & 0.797 & 0.812 & 0.827   \\
          & GC-RL & 0.945 & 0.986 & 0.774 & 0.852 & 0.836 & 0.879  \\
    \midrule
    \multirow{2}[2]{*}{Qwen2.5-3B} & C-RL  & 0.752 & 0.754 & 0.652 & 0.699 & 0.691 & 0.710   \\
          & GC-RL & 0.774 & 0.802 & 0.657 & 0.712 & 0.698 & 0.729  \\
    \midrule
    \multirow{2}[2]{*}{Qwen2.5-1.5B} & C-RL  & 0.651 & 0.638 & 0.607 & 0.648 & 0.612 & 0.631   \\
          & GC-RL & 0.683 & 0.676 & 0.658 & 0.672 & 0.635 & 0.665  \\
    \bottomrule
    \end{tabular}%
      \caption{Supplemental critique capability evaluation results on Qwen-2.5-(1.5B,3B,7B). The average denoised reward (\S \ref{subsec:denoise}) is also reported to reflect the critique capability.}
  \label{tab:denoised_reward_eval}%
\end{table*}%

\end{document}